\documentclass[sigconf]{acmart}

\usepackage{booktabs} 

\setcopyright{rightsretained}

\usepackage{blindtext}
\usepackage{graphicx}
\usepackage{hyperref}
\usepackage{tikz}
\usepackage{amsmath}
\usepackage{algpseudocode}
\usepackage{graphicx}
\usepackage{float}
\usepackage{adjustbox}
\usepackage{multirow}
\usepackage{subcaption}
\usepackage{booktabs}
\usepackage{lscape}
\usepackage{algorithm}
\usepackage{color,soul}
\usepackage{tabu}
\usepackage{array}
\newcolumntype{P}[1]{>{\centering\arraybackslash}p{#1}}
\begin{document}

\title{Self-adaptation of Genetic Operators Through Genetic Programming Techniques}

\copyrightyear{2017} 
\acmYear{2017} 
\setcopyright{acmlicensed}
\acmConference{GECCO '17}{July 15-19, 2017}{Berlin, Germany}\acmPrice{15.00}\acmDOI{http://dx.doi.org/10.1145/3071178.3071214}
\acmISBN{978-1-4503-4920-8/17/07}

\author{Andres Felipe Cruz-Salinas}
\affiliation{%
  \institution{Universidad Nacional de Colombia, Artificial Life Research Group (Alife)}
}
\email{afcruzs@unal.edu.co}

\author{Jonatan Gomez Perdomo}
\affiliation{%
  \institution{Universidad Nacional de Colombia, Artificial Life Research Group (Alife)}
}
\email{jgomezpe@unal.edu.co}

\begin{abstract}

Here we propose an evolutionary algorithm that self modifies its operators at the same time that candidate solutions are evolved. This tackles convergence and lack of diversity issues, leading to better solutions. Operators are represented as trees and are evolved using genetic programming (GP) techniques. The proposed approach is tested with real benchmark functions and an analysis of operator evolution is provided.

\end{abstract}



\maketitle

\section{Introduction}

Evolutionary algorithms (EAs) are metaheuristic optimization techniques inspired in biological evolution. A population of candidate solutions is maintained on each generation, and every candidate solution is encoded in an appropriate space in order to apply bio-inspired operators like selection, reproduction and mutation. A fitness function is defined in order to measure the quality of individuals. EAs present some issues that affect their performance: parameter tuning, premature convergence and lack of diversity.

Manual parameter tuning is the process of manually assigning parameter values to an EA. This process is, in general, tedious and time consuming. Parameter adaptation avoids the manual parameter tuning process and instead, values are modified by the algorithm according to certain rules that are predefined. For example, the one fifth rule controls the strength of the mutation according to its previous success\cite{schwefel}. In general, a static set of rules may work well in some kind of problems but are not general enough to work on other kind of problems. 

Premature convergence is an issue that arises in population based strategies. Due to the pressure to obtain solutions that optimize a given problem, individuals converge quickly to local optima. An ad-hoc strategy is to increase the population size, but in a high dimensional problem it may cause a huge overhead. Another strategy is to force diversity in different ways, relax the pressure scheme or include new randomly-generated individuals. Those strategies compromise the population quality and may help to increase diversity, but the improvement on the best individual may not be significant. Finally, crowding and niching techniques make the population converge to different local optima at the same time, recombining individuals with similar mates in order to perform exploitation in different areas of the problem space.

Self-modifying operators attempt to tackle those issues (parameter selection, premature convergence and the lack of diversity) by changing the way individuals are generated according to the current population. Self-modification also provides individuals with higher quality due to a better exploration of the problem space. 

Our proposal is an evolutionary algorithm where operators are defined as GP trees and are subject to evolution at the same time candidate solutions are evolved. This includes an additional source of diversity because the way the individuals is transformed changes along the algorithm execution. As a result of this diversity increase, the convergence of the algorithm is delayed and it leads to better results.

\section{Previous work}

In general, EAs use a fixed set of operators to be applied while evolving candidate solutions. These operators are inspired in biological evolution processes like reproduction of organisms, and have parameters that are usually tuned before running the algorithm\cite{mitchellga}.

There has been extensive work on self adapting the parameters at the same time the optimization process is carried on, especially in the continuous domain\cite{kramer}. One of the most successful methods for continuous optimization is the CMA-ES \cite{hanser} (Covariance Matrix Adaptation - Evolution Strategy), a widely applied strategy to solve real optimization problems which are non-linear and non-convex, especially when the objective function is ill-conditioned. There is also some research in self adapting parameters for combinatorial problems \cite{younes} \cite{maruo}. Most of the work is focused on tuning numerical parameters of operators like mutation rates, and crossover points.

Recently, self adaptation has been done in specific types of problems, for example, a recent approach in \cite{coelho2016hybrid} self-adapts the mutation operators guiding the search into the solution space using a self-adaptive reduced variable neighborhood search procedure in combinatorial problems. Another approach to solve multiobjective problems using self-adaptation is described in \cite{hadka2013borg}.

Finally, the approach described here is similar to ADF (automatically defined functions) proposed in \cite{koza1994genetic} and \cite{koza1996use}. As defined in \cite{koza1996use}: An automatically defined function (ADF) is a function that is dynamically evolved during a run of genetic programming and that may be called by a calling program (or subprogram) that is concurrently being evolved. The main difference with the proposed approach, is that the evolved operators are meant to transform the individuals during the algorithm execution, whereas an ADF acts as reusable components that may be called by evolved programs in a genetic programming algorithm.

\subsection{Parameter tuning}

Manual parameter tuning is one of the most important aspects to consider in EAs. Due to the no free lunch theorem \cite{nofreelunch}, there are no unique parameter values or even a unique algorithm that performs equally well for all optimization problems. An ad-hoc strategy is to perform the process manually, but it can be expensive and it is problem-dependent. This leads to automatic parameter tuning, where the researcher allows the algorithm to test multiple configurations of parameters and choose the one that works best. The F-Race and iterated F-Race methods \cite{birattari2010f} use statistical information for selecting the best configuration out of a set of candidate configurations under stochastic evaluations. In \cite{lopez2016irace} there is a description of these methods as well as some improvements to the iterated racing method implemented in a software package called \textit{irace}. Either manual or automatic, parameter tuning remains expensive, because the EA must be run in order to measure the effectiveness of a given configuration.

\subsection{Parameter adaptation}

Parameter adaptation (or control) is a strategy to find good parameter values without doing a manual search on every problem. The approach is to modify the parameter values at the same time the EA is searching for solutions. The ways to adapt the parameters are broad. There is an overview of techniques applied to numerical problems in \cite{agoston}. A common strategy is related to the mutation rate, and another common approach described in \cite{agoston}, is to adapt through time the penalty of the fitness function, this is related to constrained optimization problems. Finally, \cite{agoston} does a classification of the adaptation strategies in three categories: deterministic, adaptive and self-adaptive. 

\subsection{Adaptation through operator rates}

There is a strategy that considers an EA in a higher level, it is to apply a single operator (from an operator set) on every generation depending on how good the operator is. This is achieved by associating an operator to a rate that measures its quality. In \cite{gomezhaea}, there is a brief description of two rough categories of this approach: centralized and decentralized techniques.

Finally, \cite{gomezhaea} proposes a hybrid approach by evolving the operator rates without using special metaoperators, the probability of choosing an operator at every iteration is either ``punished" (decreased) or ``rewarded" (increased), depending on the improvement of the individual when the operator is applied. As in decentralized techniques, the rates are normalized in order to sum one. The selection of the operator is a typical procedure (e.g., roulette or tournament) using the rates as the operator fitness. 

\section{Proposed approach}
The aim of this work is to go further from solely self-adapting the parameters in EAs towards self-modifying the structure of operators using GP techniques. This approach is a generalized version of HAEA \cite{gomezhaea}: the operators belong to an operators population and are exposed to evolution (like a coevolutionary technique). The strategy to select the operators is still the hybrid approach of \cite{gomezhaea}, where a typical selection method (roulette or tournament) picks an operator proportionally to its probability to be chosen. From now on, AOEA (Adaptive Operators Evolutionary Algorithm) will refer to the proposed approach.

\begin{figure}
	\includegraphics[scale=0.35]{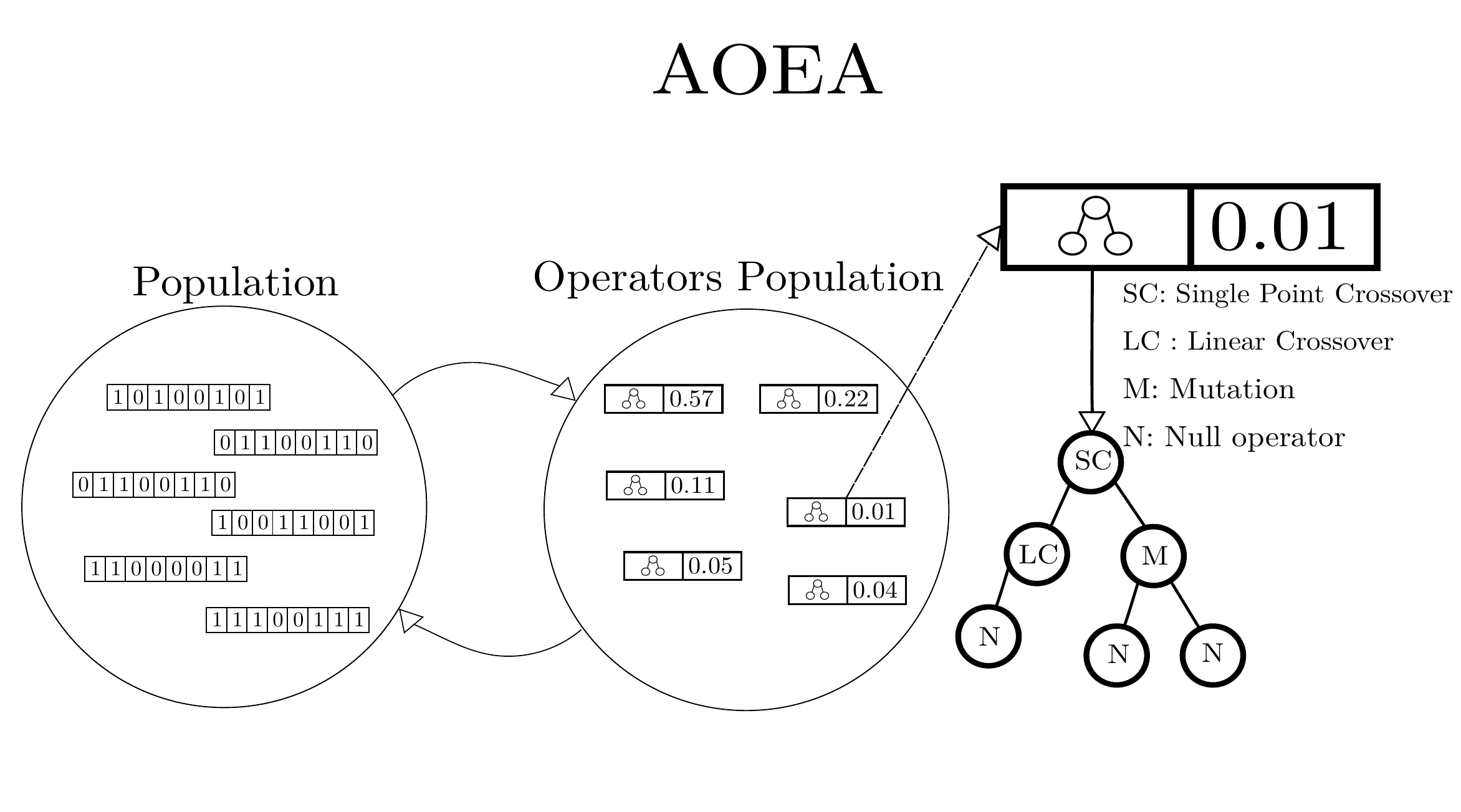} \hfill
	\caption{Graphical representation of the populations. In this case, the candidate solution encoding is binary. Each operator has two elements: A tree structure, and its probability to be chosen.}
\end{figure}

\subsection{Operators as genetic programming trees}

Here, the approach is to change the static structure of an operator and convert it into a genetic programming tree.

\textbf{Atomic operator}: An atomic operator is predefined by the user and is not exposed to evolution. It is one dimensional if defined as $o : D \rightarrow D$ where $D$ is the search space of the problem being solved. Similarly, an operator is said to be two dimensional if defined as $o : D^{2} \rightarrow D$. These operators capture the notion of mutation and crossover (respectively) in traditional genetic algorithms. A 1D operator takes a single individual as an argument and returns a modified version based on it (like classical mutation). Similarly, a 2D operator takes two individuals and produces a single individual, a ``child" (this operation is not necessarily commutative). There is a special type of 2D atomic operator, called the \textit{null} operator, which always return either the first or the second individual without modification.

\textbf{Operator}: An operator is defined as a binary tree, where each node contains either a 1D or a 2D atomic operator. The operator always receives two individuals as arguments, but it is up to the arity of the atomic operators the possible recombination of individuals. In order to compute the full transformation of an operator, a post-order traversal \cite{clrs} is performed applying atomic operators transformations in a top-down fashion. A node performs an atomic operation with the arguments equal to the result of its children operators. This process continues recursively until a leaf is reached, in which case one of the two arguments is returned without modification (the \textit{null} operator). The choice of which individual its returned is deterministic.

Formally, the operator $O$ is defined as a triple:
\[
	O = [O_l, O_r, o]
\]

Where $O_l$ is the left child of $O$, $O_r$ is the right child of $O$ and $o$ is an atomic operator. The result of an operator is computed as shown in equation \ref{operator}

\begin{equation} 
  O(A,B)=\begin{cases}
    o( O_{l}(A,B), O_{r}(A,B) ), & \text{if $ \tau(O) = 2 $}.\\
    o( O_{r}(A,B) ), & \text{if $ \tau(O) = 1 $}.\\
    o(A,B), & \text{otherwise}.
  \end{cases}
  \label{operator}
\end{equation}

$\tau(O)$ defines the number of children of a given node $O$. As in genetic programming, these trees are subject to mutation and recombination. Here, we use the following operators:

\begin{itemize}  
\item \textbf{Mutation:} A mutation occurs on a random node of the tree, and the atomic operator of that node is randomly changed by an atomic operator of the same arity.

\item \textbf{Recombination:} This meta operator takes two trees, selects a random node from each one, then swaps the subtrees from which the chosen nodes are roots. This method is described in \cite{koza1992genetic}.
\end{itemize}

\emph{Random node selection:} In order to support the above operations a random node procedure must be defined. A reservoir sampling technique is applied in order to return a uniformly distributed node from the given tree.

The random operators population and the recombination/mutation procedures always guarantee that the number of children for a given node is consistent with the cardinality of its atomic operator. It is also guaranteed that the leaves always contain a \textit{null} atomic operator. In the figure \ref{fig:treeComposition}, there is a graphical example of an operator and how the atomic operators are composed in order to form the full operator.

\begin{figure}

\begin{tikzpicture}[auto, node distance=1.8cm, every loop/.style={},
                    thick,main node/.style={circle,draw,font=\sffamily\Large\bfseries}]

  \node[main node] (1) {$O_{0}, f_{0}$};
  \node[main node] (2) [below left of=1] {$O_{1}, f_{1}$};
  \node[main node] (3) [below right of=1] {$O_{2}, g_{0}$};
  \node[main node] (4) [below left of=2] {$O_{3}, h_{3}$};
  \node[main node] (5) [below right of=2] {$O_{4}, h_{4}$};
  \node[main node] (6) [below right of=3] {$O_{5}, h_{5}$};

  \path[every node/.style={font=\sffamily\small}]
    (1) edge node [left] {} (2)
	(1) edge node [right] {} (3)    
    (2) edge node [left] {} (4)
	(2) edge node [right] {} (5)    
    (3) edge node [right] {} (6)
    ;
\end{tikzpicture}

\[O_{0}(A,B) = f_{0}( O_{1}(A,B), O_{2}(A,B) )\]
\[O_{1}(A,B) = f_{1}( O_{3}(A,B), O_{4}(A,B) )\]
\[O_{3}(A,B) = h_{3}(A,B) = A\]
\[O_{4}(A,B) = h_{4}(A,B) = B\]
\[O_{2}(A,B) = g_o(O_5(A,B))\]
\[O_{5}(A,B) = h_{5}(A,B) = B\]	

\caption{Tree representation of an operator and its unfolded composition. The first element in each node represents its label, and the second element represents the atomic operator that belongs to that node. Below there is the operation performed on each node according to their children and their atomic operators. The atomic operators, labelled as $f$, $g$ and $h$ represent a two dimensional operator, a one dimensional operator, and the \textit{null} operator, respectively.}
\label{fig:treeComposition}
\end{figure}

\subsection{Punish reward scheme}

Based on HaEa\cite{gomezhaea}, a punish reward scheme is defined in order to evolve individuals according to the operators quality. A operators population is defined with size $\kappa$. An operator's quality is represented by a number from the range $[0,1]$ (a rate), this measure is increased or decreased according to the performance of the operator on each generation. At the beginning of the algorithm, the quality measure is set to a random value. The operators population is evolved on each generation using the described recombination and mutation methods for genetic programming trees, and the selection is proportional to the quality measure previously described. The outline of the algorithm is presented in Algorithm \ref{alg:aoea}.

\begin{algorithm}
\begin{algorithmic}[1]
\Function{AOEA}{$\lambda$,$\kappa$}
\State{$t_{0} = 0$}
\State{$P_{0} = initPopulation(\lambda)$}
\State{$O_{0} = initOperators(\kappa)$}
\State{$R_{0} = initRates(\kappa)$}

\While{not terminationCondition($P_{t}$)}
	\State{crossoverPopulation($P_{t}$,$O_{t}$, $R_{t}$)}
	\State{crossoverOperators($P_{t}$,$O_{t}$,$\kappa$)}
	\State{mutationOperators($P_{t}$,$O_{t}, \kappa$)}
	\State{$t = t + 1$}
\EndWhile
\State \Return best(P)
\EndFunction

\end{algorithmic}
\caption{AOEA outline. A Java implementation of this algorithm can be founded here \url{https://github.com/afcruzs/AOEA}.}
\label{alg:aoea}
\end{algorithm}

On every iteration of the algorithm, an individual randomly selects one operator proportionally to the operators quality, and another individual is selected proportionally to its fitness value. These two individuals are recombined with the previously chosen operator, and the parent is replaced with its child if and only if the child fitness is equal or better. For each operator, there will be a vote system in order to measure the operator quality. If the fitness is better, there will be a positive vote, if it is worse, there will be a negative vote. This procedure is repeated for every individual in the population. Then, an operator will be rewarded if its vote count is positive, punished if it is negative. If it is zero, there will be no changes. After the rates are modified they are normalized.

The selection of individuals and operators is performed using a roulette selection algorithm. This recombination process is described in the algorithm \ref{alg: crossover_population}.\newline

\begin{algorithm}
\begin{algorithmic}[1]
\Procedure{crossoverPopulation}{$P_{t}$,$O_{t}$, $R_{t}$}
        	\State{	$P_{t+1} = \{\}$ }
        	\State{	$R_{t+1} = R_{t}$ }        	
        	\State{	$V = initVotesToZero(\kappa)$ }            	
        	\For{ \textbf{each} ind $\in $ $P_{t}$  }    
                \State{$operator = selectOperator(O_{t}, R_{t})$ }
                \State{$mate = selectIndividual(P_{t})$ }
                \State{$child1 = operator(ind,mate)$ }
                \State{$child2 = operator(mate,ind)$ }
                \State{$child = Best(child1,child2)$ }
                \If{fitness(child) $>$ fitness(ind)}
                    \State{$V[operator] = V[operator] + 1$}
                \Else
                	\State{$V[operator] = V[operator] - 1$}
                \EndIf
                                
        		\If{fitness(child) $>=$ fitness(ind)}
                    \State{$P_{t+1} = P_{t+1} \cup \{child\}$}
                \Else
                	\State{$P_{t+1} = P_{t+1} \cup \{ind\}$}
                \EndIf      
        	\EndFor	
        	\For{ \textbf{each} operator $\in $ $O_{t}$  }  
                \State{$\delta = random(0,1)$ }        	
        		\If{V[operator] $>$ 0}
        			\State{$R_{t+1}[operator] = (1 + \delta) * R[operator]$}
        		\ElsIf{V[operator] $<$ 0}
					\State{$R_{t+1}[operator] = (1 - \delta) * R[operator]$}
        		\EndIf
        		
				\State{normalizeRates(opRates)}   
        	\EndFor
        \EndProcedure

\end{algorithmic}
\caption{Candidate solutions crossover}
\label{alg: crossover_population}
\end{algorithm}

Then, the operators are recombined in order to evolve them proportionally to their quality using a roulette selection algorithm. Finally, a mutation on each tree is performed with probability equal to $1/\kappa$ as described above. These two processes are described in \ref{alg: crossover_operators} and \ref{alg: mutation_operators}.

\begin{algorithm}
\begin{algorithmic}[1]
\Procedure{crossoverOperators}{$O_{t}$,$R_{t}$,$\kappa$}
	\State{$O_{t+1} = \{\}$}
	\State{$shuffle(O_{t})$}	
	\For{ \textbf{each} $i$ where $i < \kappa$ and $i$ is even  }    
			\State{$mate1 = O_{t,i}$}
			\State{$mate2 = O_{t,i + 1}$}			
			\State{$child1, child2 = recombine(mate1,mate2)$}
			\State{$O_{t+1} = O_{t+1} \cup \{child1, child2\}$}
	\EndFor
\EndProcedure
\end{algorithmic}
\caption{Operators crossover}
\label{alg: crossover_operators}
\end{algorithm}

\begin{algorithm}
\begin{algorithmic}[1]
\Procedure{mutationOperators}{$P_{t}$,$O_{t}$}
	\State{$prob = 1.0 / \kappa$}
	\For{ \textbf{each} operator $\in $ $O_{t}$  }    
		\If{random(0,1) $<=$ prob}
			\State{mutateOperator(operator)}
		\EndIf
	\EndFor
\EndProcedure
\end{algorithmic}
\caption{Operators mutation}
\label{alg: mutation_operators}
\end{algorithm}

\subsection{Operators initialization}

The operators population is randomly generated before the execution of the algorithm, and in order to avoid too complex operators at the beginning every operator has a maximum depth of four. On each node, an atomic operator is chosen such that its arity is equal to the number of children. When the process reaches a leaf, a boolean flag is randomly generated to always return either the first or the second argument.

\section{Results}

The proposed approach was tested with benchmark functions shown in table \ref{benchmark-table}. These functions were selected because are standard on the real optimization literature, specially on evolutionary approaches. Each experiment is performed with 500 iterations, and with 50 and 100 individuals in the population. Additionally, the operators population $\kappa$ is fixed to 16 on every experiment. The dimension for every experiment is set to 1000 unless the function is defined with a specific dimensionality. Every experiment is repeated 50 times. Finally, the initial population is randomly generated without violating the constraints of the function and it is the same for every experiment. The chosen coding method is a simple vector of real numbers.

\begin{table*}

\centering
\caption{Benchmark real functions. Every function has an optimal value of 0.0 except for H1 (is a maximizing function) which is 2. The functions are sorted in increasing ``hardness". In general, the higher the dimension, the harder the function. On functions with the same dimension our measure of hardness is given by experimental results on how close are the results to the global optimum. The first column maps each function to an id to reference the functions on the results tables.}
\begin{tabu}{|l|l|l|l|}
\hline
\textbf{Id} & \textbf{Name} & \textbf{Function} & \textbf{Interval} \\ \hline
1 & Jong 1                & $\sum_{i=1}^{N} x_{i}^{2}$ & $x_{i} \in [-5.12, 5.12]$  \\ \hline
2 & Jong 2                & $\sum_{i=1}^{N} (i+1) x_{i}^{2}$ &  $x_{i} \in [-5.12, 5.12]$    \\ \hline
3 & Jong 3                & $\sum_{i=1}^{N} x_{i}^{i}$ & $x_{i} \in [-1, 1]$  \\ \hline
4 & Himmelblau            & $(x_{1}^{2} + x_{2} - 11)^{2} + (x_{1} + x_{2}^{2} - 7)^{2}$  & $x_{i} \in [-6, 6]$  \\ \hline
5 & \begin{tabular}[c]{@{}l@{}}Two\\ peak trap\end{tabular} & 
$ f(x) = \left\{
  \begin{array}{ll}
    \frac{160}{15}(15-x) & \mbox{if } 10 \leq x < 15 \\ \\
    \frac{200}{5}(x-15) & \mbox{otherwise} \\
  \end{array}
\right.$  & $x_{i} \in [-15, 15]$  \\ \hline

6 & \begin{tabular}[c]{@{}l@{}}Central two\\ peak trap\end{tabular} & 
$ f(x) = \left\{
  \begin{array}{ll}
    \frac{160}{15}x  & \mbox{if } x < 10 \\ \\
    \frac{160}{15}(15-x) & \mbox{if } 10 \leq x < 15 \\ \\
    \frac{200}{5}(x-15) & \mbox{otherwise} \\
  \end{array}
\right.$
 & $x_{i} \in [-15, 15]$  \\ \hline
 
7 &  H1     & $\frac{sin(x_{1} - \frac{x_{2}}{8})^{2} + sin(x_{2} + \frac{x_{1}}{8})^{2}}{\sqrt{(x_{1} - 8.6998)^{2} + (x_{2} - 6.7665)^{2} + 1}}$  & $x_{i} \in [-100, 100]$  \\ \hline

8 & Ackley & 
$\begin{array}{l}
20 - 20 * exp(-0.2 * \sqrt{\frac{1}{N} \sum_{i=1}^{N} x_{i}^{2}}) + \\ e - exp(\frac{1}{N} \sum_{i=1}^{N} cos(2\pi x_{i})) \end{array}$  
& $x_{i} \in [-5, 5]$  \\ \hline
9 & Shubert 2D & $\begin{array}{l}(\sum_{i=1}^{5} cos((i+1) * x_{0} + i))* \\ (\sum_{i=1}^{5} cos((i+1) * x_{1} + i)) \end{array}$ & $x_{i} \in [-5.12, 5.12]$ \\ \hline
10 & Griewangk             & $ \frac{1}{4000} \sum_{i=1}^{N} x_{i}^{2} - \prod_{i=1}^{N} cos(\frac{x_{i}}{\sqrt{i}}) + 1 $ & $x_{i} \in [-600, 600]$ \\ \hline

11 & Rastrigin & $10 * N + \sum_{i=1}^{N} x_{i}^2 - 10 * cos(2\pi x_{i})$ & $x_{i} \in [-5.12, 5.12]$  \\ \hline

12 & Schaffer & $\begin{array}{l}\sum_{i=1}^{N-1} (x_{i}^{2} + x_{i+1}^{2})^{0.25} * [sin(50* \\ (x_{i}^{2} + x_{i+1}^{2})^{0.1}) + 1] \end{array}$   & $x_{i} \in [-100, 100]$     \\ \hline

13 & Rosenbrock & $\sum_{i=1}^{N-1} 100*(x_{i+1} - x{i}^2) + (1-x_{i})^2$ & $x_{i} \in [-2.048, 2.048]$  \\ \hline

14 & Bohachevsky           & $\begin{array}{c} \sum_{i=1}^{N-1} (x_{i}^{2} + 2x_{i+1}^{2} -0.3cos(3 \pi x_{i}) - \\ 0.4 cos(4 \pi x_{i+1}) + 0.7) \end{array}$ & $x_{i} \in [-100, 100]$  \\ \hline

15 & Schwefel & $418.9829 * n \sum_{i=1}^{n} x_{i} * sin(\sqrt{|x_{i}|})$ & $x_{i} \in [-500, 500]$  \\ \hline

\end{tabu}

\label{benchmark-table}
\end{table*}

The numerical results of the experiments are shown in tables \ref{results-50} and \ref{results-100}. besides the proposed approach (AOEA), there is GA (genetic algorithm) and HAEA (Hybrid adaptive evolutionary algorithm from \cite{gomezhaea}). Those were implemented in order to have a baseline for comparison. In the GA the recombination method is linear crossover with random weights, the mutation operator is gaussian noise added to a random position of the chromosome. Finally, HAEA does not have parameters to tune besides the population size. For each iteration the best individual of each experiment is stored in order to visualize the convergence and compare the algorithms.

The atomic operators used in this experiment (for both HAEA and AOEA) are the following:
\begin{itemize}
\item swap two randomly chosen genes,
\item add gaussian noise to a randomly chosen gene,
\item single point crossover,
\item uniform crossover,
\item average crossover,
\item linear crossover.
\end{itemize}

\begin{figure}
	\centering 
	\includegraphics[scale=0.225]{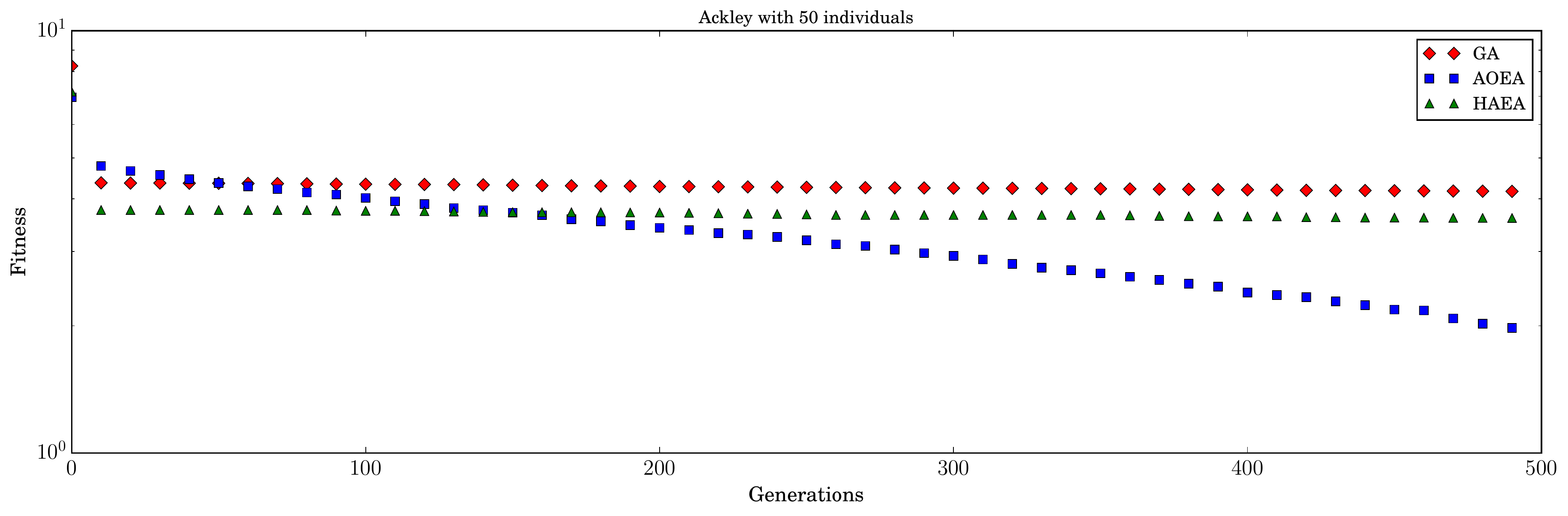} \hfill
	\includegraphics[scale=0.225]{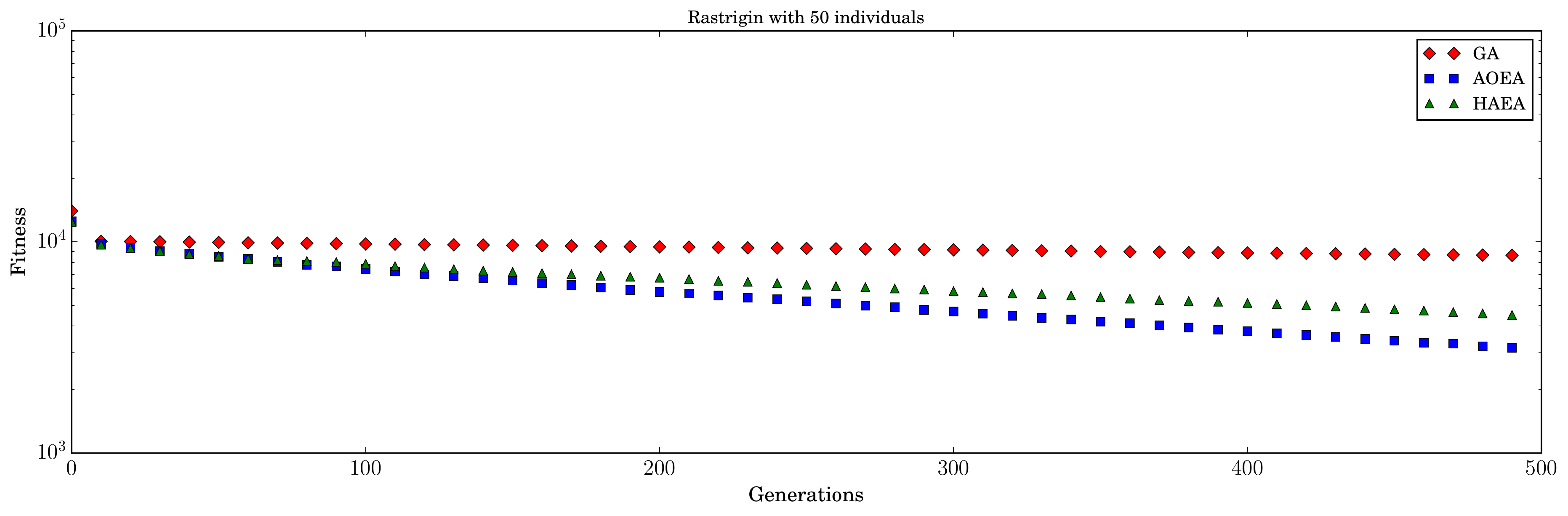} \hfill
	\caption{Median of Ackley and Rastrigin functions with 50 individuals in the population. The fitness is on logarithmic scale.}
	\label{hard-1}
\end{figure}

\begin{figure}
	\centering 
	\includegraphics[scale=0.225]{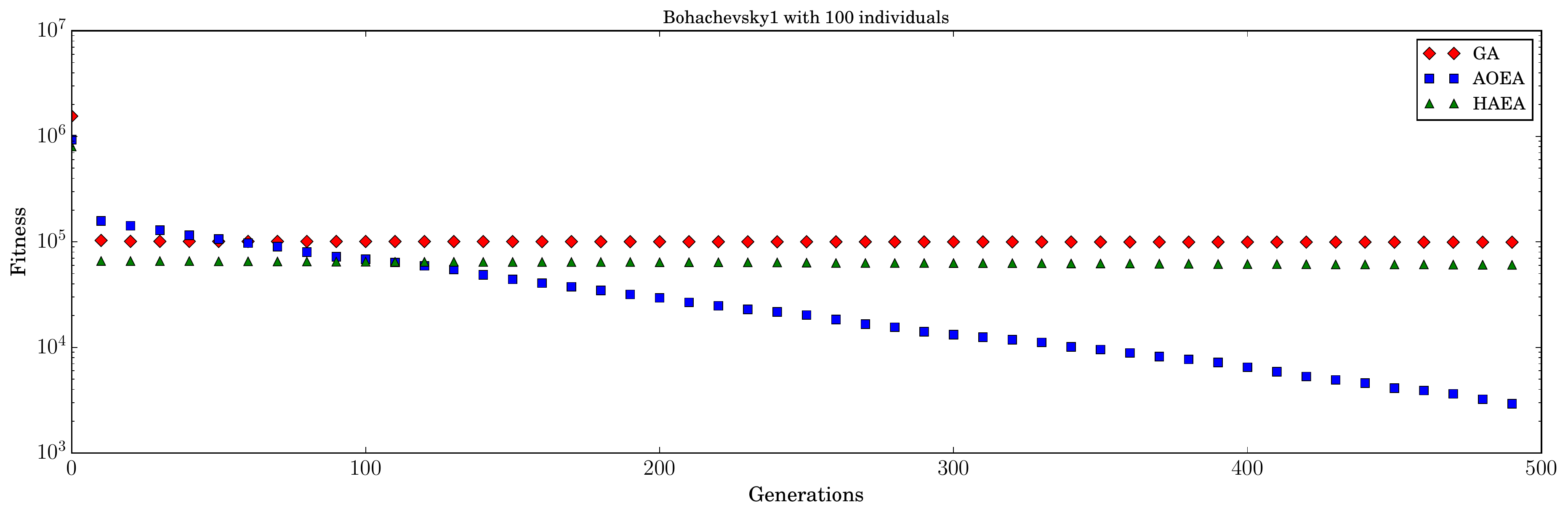} \hfill
	\includegraphics[scale=0.225]{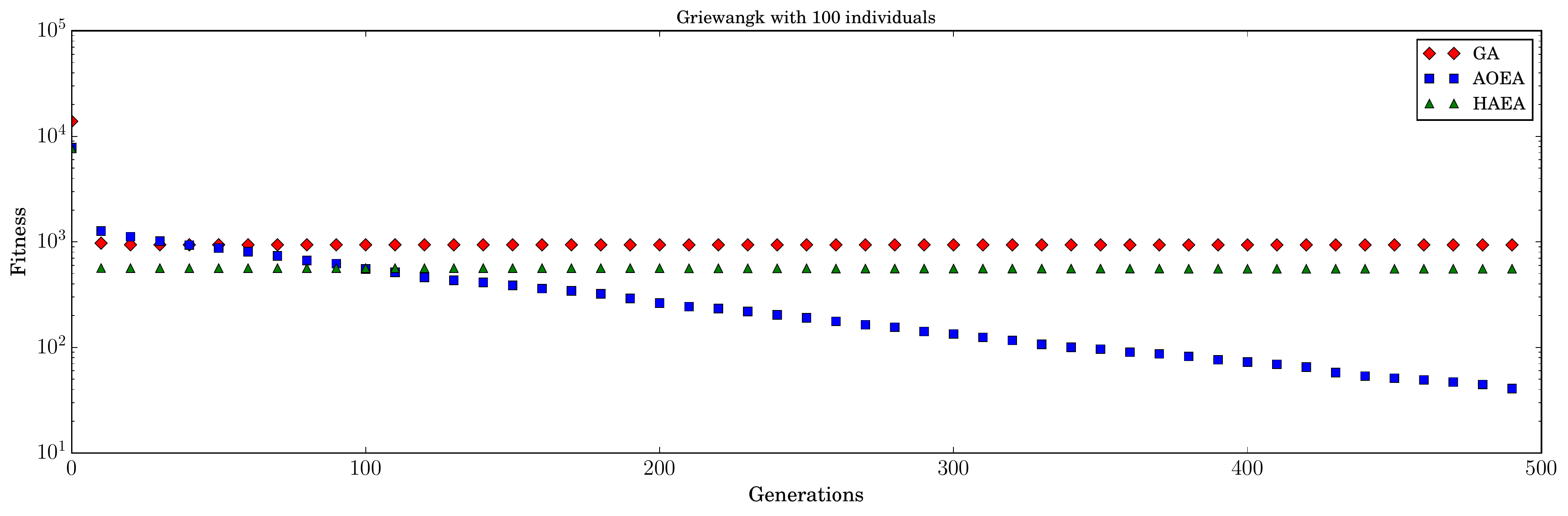} \hfill
	\caption{Median of Bohachevsky and Griewangk functions with 50 individuals in the population. The fitness is on logarithmic scale.}
	\label{hard-2}	
\end{figure}

\begin{figure}
	\centering 
	\includegraphics[scale=0.225]{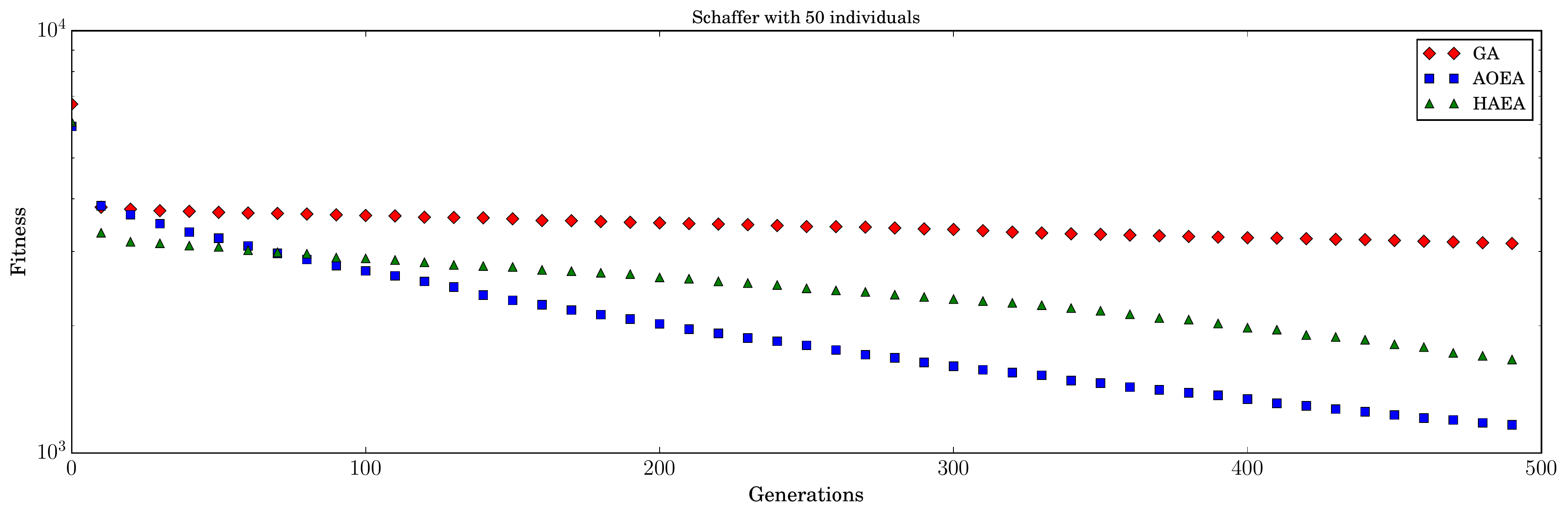} \hfill
	\includegraphics[scale=0.225]{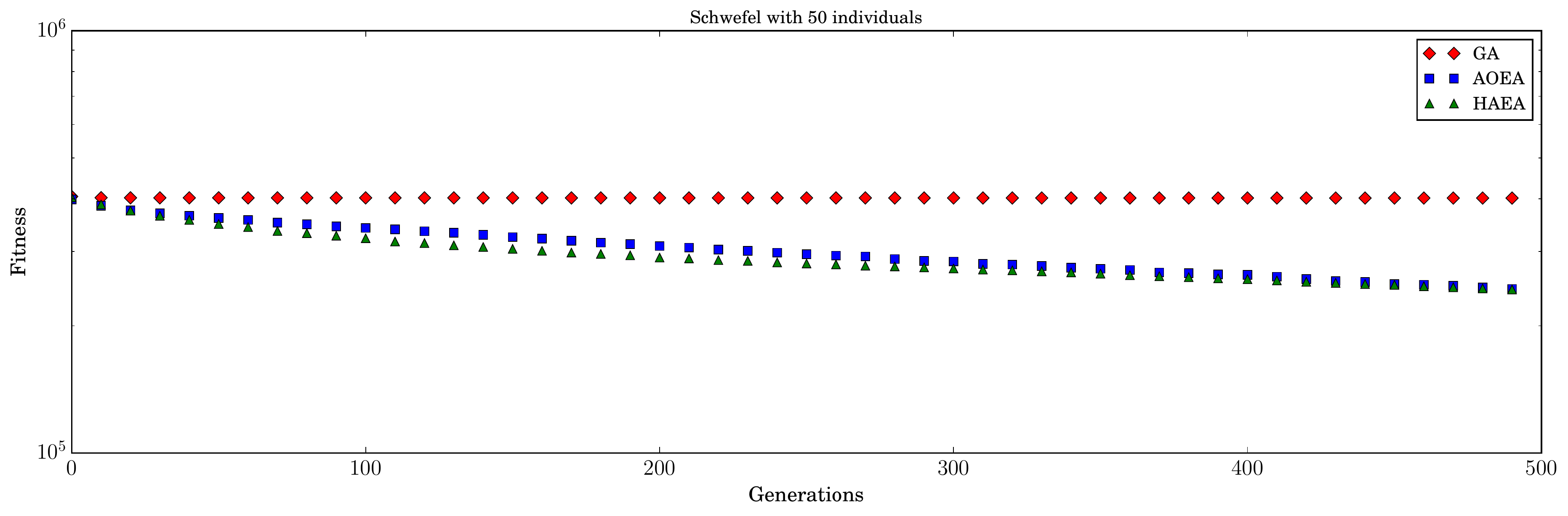} \hfill
	\caption{Median of Schaffer and Schwefel functions with 50 individuals in the population. The fitness is on logarithmic scale.}
	\label{hard-3}	
\end{figure}

\begin{figure}
	\begin{subfigure}{\linewidth}
		\includegraphics[scale=0.225]{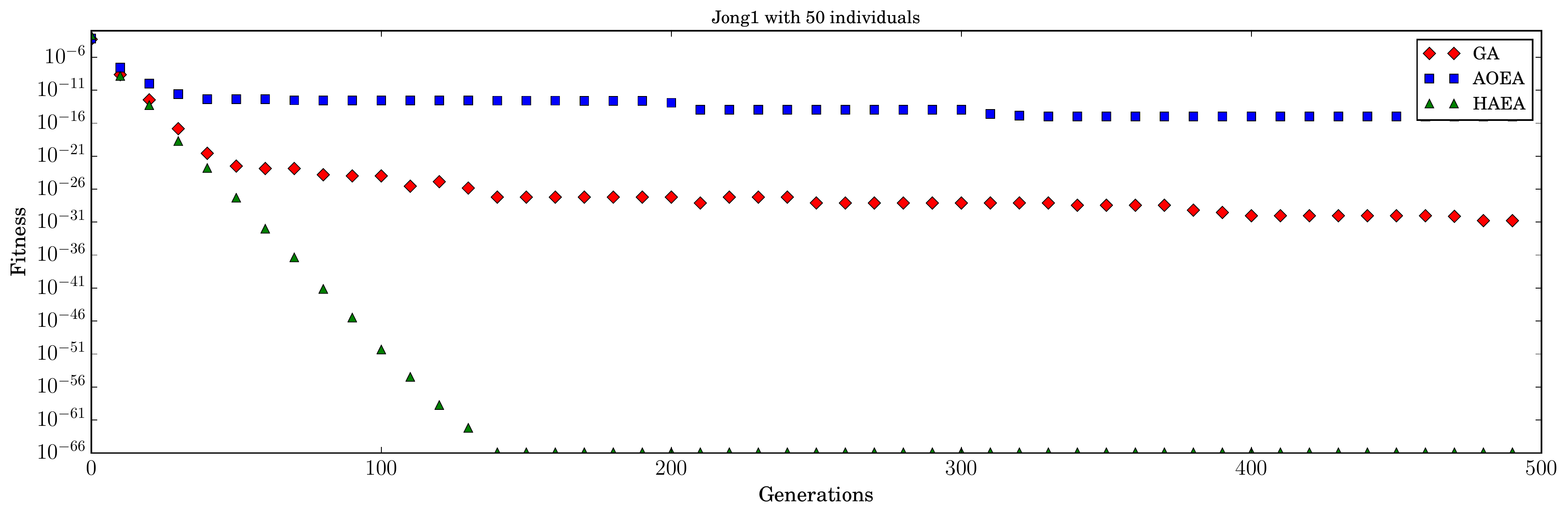}
		\includegraphics[scale=0.225]{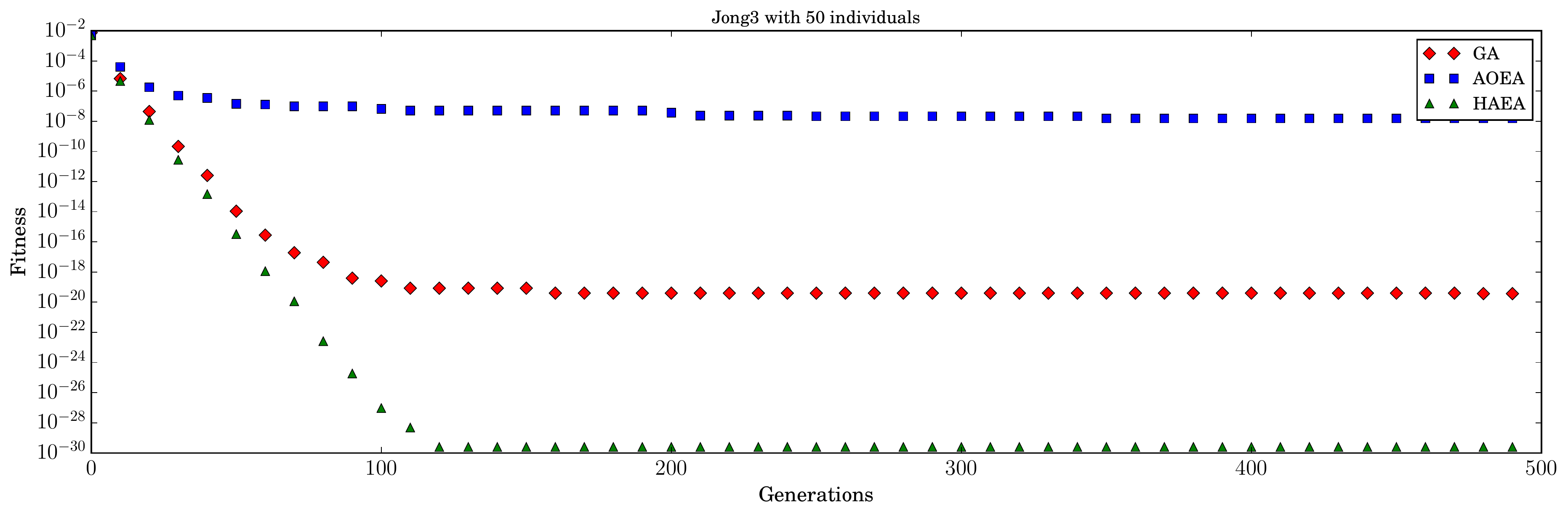} 
		\includegraphics[scale=0.225]{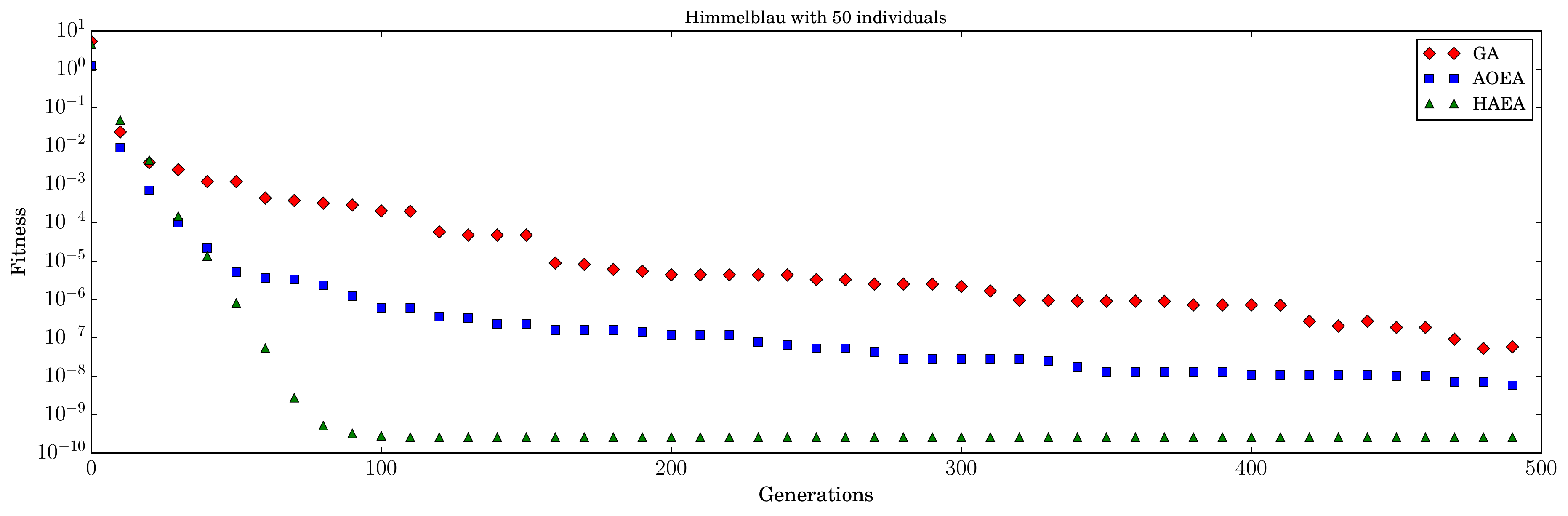} 
		\includegraphics[scale=0.225]{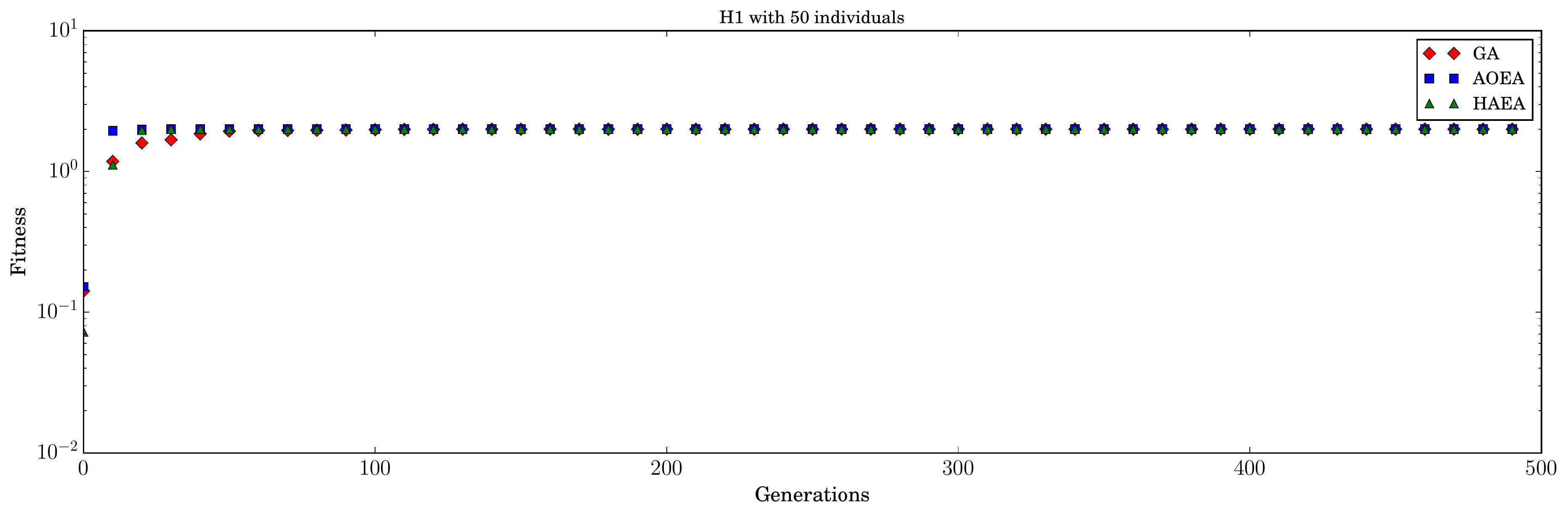} 	
	\end{subfigure}
	\caption{Functions with low dimensionality: Median of Jong1, Jong3, Himmelbau and H1. With 50 individuals in the population. The fitness is on logarithmic scale.}
	\label{easy-1}	
\end{figure}

\begin{table}
\caption{Results of the last generation with 50 individuals in the population. Best results are in bold.}
\begin{tabular}{|l|l|l|l|}
\hline
           \textbf{Id} &             \textbf{GA median} &            \textbf{HAEA median} &             \textbf{AOEA median} \\ \hline
	1  &      1.6E-31 $\pm$ 0.0 &       1.4E-66 $\pm$ 0.0 &      \textbf{5.4E-70 $\pm$ 0.0} \\ \hline
	2  &      1.9E-35 $\pm$ 0.0 &       5.7E-78 $\pm$ 0.0 &      \textbf{2.3E-81 $\pm$ 0.0} \\ \hline
	3  &      3.6E-20 $\pm$ 0.0 &       2.4E-30 $\pm$ 0.0 &      \textbf{9.0E-45 $\pm$ 0.0} \\ \hline
	4  &      5.8E-08 $\pm$ 0.0 &       2.5E-10 $\pm$ 0.0 &      \textbf{1.1E-11 $\pm$ 0.0} \\ \hline
	5  &      \textbf{0.0E+00 $\pm$ 0.0} &       5.3E-03 $\pm$ 0.010 &      1.7E-04 $\pm$ 0.0 \\ \hline
	6  &      \textbf{0.0E+00 $\pm$ 0.0}  &       4.1E-03 $\pm$ 0.010 &      9.6E-05 $\pm$ 0.0 \\ \hline	
	7  &      \textbf{2.0E+00 $\pm$ 0.3} &       \textbf{2.0E+00 $\pm$ 0.001} &      \textbf{2.0E+00 $\pm$ 0.0} \\ \hline
	8  &      4.1 $\pm$  0.1 &       3.5 $\pm$ 0.3 &      \textbf{7.8E-01 $\pm$ 0.185} \\ \hline 
	9  &     \textbf{1.8E+02 $\pm$ 21.6} &       \textbf{1.8E+02 $\pm$ 0.0} &      \textbf{1.8E+02 $\pm$ 0.0} \\ \hline
	10 &     1.8E+03 $\pm$ 279.5 &    1.0E+03 $\pm$ 1.3E+03 &      \textbf{2.2E+01 $\pm$ 8.2} \\ \hline
	11 &     8.6E+03 $\pm$ 152.4 &     4.4E+03 $\pm$ 6E+02 &    \textbf{1.4E+03 $\pm$ 167.6} \\ \hline
	12 &     3.1E+03 $\pm$ 166.2 &     1.6E+03 $\pm$ 283.3 &     \textbf{1.1E+03 $\pm$ 57.1} \\ \hline
	13 &     1.0E+04 $\pm$ 1.2E+03 &    6.2E+03 $\pm$ 7.1E+03 &     \textbf{1.2E+03 $\pm$ 78.923} \\ \hline
    14 &     1.8E+05 $\pm$ 2.5E+04 &  1.0E+05 $\pm$ 1.6E+05 &   \textbf{7.5E+02 $\pm$ 1282.9} \\ \hline
	15 &     4.0E+05 $\pm$ 3220.7 &    2.4E+05 $\pm$ 7648.8 &  \textbf{8.5E+04 $\pm$ 1.7E+04} \\ \hline          

\end{tabular}
\label{results-50}        
\end{table}

\begin{table}
\caption{Results of the last generation with 100 individuals in the population. Best results are in bold.}
\begin{tabular}{|l|l|l|l|} \hline
\textbf{Id} &             \textbf{GA median} &            \textbf{HAEA median} &             \textbf{AOEA median} \\ \hline
           
	1  &     1.288E-57 $\pm$ 0.0 &      \textbf{8.349E-266 $\pm$ 0.0} &     1.762E-173 $\pm$ 0.0 \\ \hline
	2  &     2.0E-57 $\pm$ 0.0 &      \textbf{1.5E-265 $\pm$ 0.0} &     1.5E-174 $\pm$ 0.0 \\ \hline
	3  &     4.0E-46 $\pm$ 0.0 &      \textbf{1.9E-138 $\pm$ 0.0} &      4.0E-92 $\pm$ 0.0 \\ \hline
	4  &     2.4E-10 $\pm$ 0.0 &      6.2E-14 $\pm$ 0.0 &       \textbf{6.0E-24 $\pm$ 0.0} \\ \hline
	5  &     \textbf{0.0E+00 $\pm$ 0.0} &      1.9E-03 $\pm$ 0.005 &      5.8E-05 $\pm$ 0.0 \\ \hline
	6  &     \textbf{0.0E+00 $\pm$ 0.0} &      3.0E-03 $\pm$ 0.006 &      5.4E-05 $\pm$ 0.0 \\ \hline
	7  &     \textbf{2.0E+00 $\pm$ 0.001} &       \textbf{2.0E+00 $\pm$ 0.0} &       \textbf{2.0E+00 $\pm$ 0.0} \\ \hline
	8  &     3.4E+00 $\pm$ 0.1 &      3.0E+00 $\pm$ 0.4 &       \textbf{4.0E-01 $\pm$ 0.0} \\ \hline
	9  &     \textbf{1.8E+02 $\pm$ 0.005} &       \textbf{1.8E+02 $\pm$ 0.0} &       \textbf{1.8E+02 $\pm$ 0.0} \\ \hline
	10 &     9.3E+02 $\pm$ 93.2 &    5.5E+02 $\pm$ 519.15 &       \textbf{8.2E+00 $\pm$ 2.2} \\ \hline
	11 &     8.2E+03 $\pm$ 146.6 &    4.3E+03 $\pm$ 956.0 &      \textbf{7.9E+02 $\pm$ 86.6} \\ \hline
	12 &     2.6E+03 $\pm$ 75.5 &    1.5E+03 $\pm$ 292.1 &      \textbf{8.7E+02 $\pm$ 48.9} \\ \hline
	13 &     5.5E+03 $\pm$ 532.9 &   3.5E+03 $\pm$ 1420.4 &      \textbf{1.058E+03 $\pm$ 17.2} \\ \hline
    14 &     9.9E+04 $\pm$ 7621.1 &  6.0E+04 $\pm$ 62534.6 &     \textbf{1.6E+02 $\pm$ 530.4} \\ \hline
	15 &     4.0E+05 $\pm$ 1635.1 &   1.817E+05 $\pm$ 7809.3 &   \textbf{3.6E+04 $\pm$ 11764.7} \\ \hline
          
\end{tabular}
\label{results-100}
\end{table}

Figures \ref{hard-1}, \ref{hard-2} and \ref{hard-3} contain some of the ``hardest" functions of the tested dataset, the candidate solutions start far from the optimal value and the dimension is higher compared to the other functions. In general, the proposed strategy converges much later than the traditional GA and HAEA giving better results. The situation is a bit different in the set of functions where the dimensionality is fixed to two (figure \ref{easy-1}): in those cases solutions are very close to an optimal value of 0.0 and those cases HAEA generally performs better than the proposed approach. Nevertheless, the solutions of almost every algorithm are very good because the analytic form of the functions is simple and the dimensionality is two compared to 1000 on the ``harder" functions.

\subsection{Statistical tests}

The proposed approach is compared to the baseline algorithms (classic GA and HAEA \cite{gomezhaea}) by applying a statistical test over the best individual (on each experiment) in the last generation on every objective function and population size. Initially, the measurements were tested using a D'Agostino's K-squared normality test. Only about 45\% of all the experiments passed the test, so a Wilcoxon signed-rank test was used with the null hypothesis that the paired samples come from the same distribution with a 95\% confidence interval.

\begin{table}
\caption{Experiments on which the null hypothesis was not rejected. AOEA vs GA.}
\centering
\begin{tabular}{|l|l|l|l|l|}
\hline
\textbf{Function} & \textbf{Population} & \textbf{\begin{tabular}[c]{@{}l@{}}Positive\\ sum\end{tabular}} & \textbf{\begin{tabular}[c]{@{}l@{}}Negative\\ sum\end{tabular}} & \textbf{W} \\ \hline
Himmelblau        & 100                 & 794.0                 & 481.0                 & 481.0      \\ \hline
Shubert2D        & 100                 & 811.0                 & 464.0                 & 464.0      \\ \hline
Shubert2D        & 50                 & 775.0                 & 500.0                 & 500.0      \\ \hline
\end{tabular}
\label{results-ga}
\end{table}

\begin{table}
\caption{Experiments on which the null hypothesis was not rejected. AOEA vs HAEA.}
\centering
\begin{tabular}{|l|l|l|l|l|}
\hline
\textbf{Function} & \textbf{Population} & \textbf{\begin{tabular}[c]{@{}l@{}}Positive\\ sum\end{tabular}} & \textbf{\begin{tabular}[c]{@{}l@{}}Negative\\ sum\end{tabular}} & \textbf{W} \\ \hline
H1        & 100                 & 587.0                 & 688.0                 & 587.0      \\ \hline
Himmelblau        & 100                 & 444.0                 & 831.0                 & 444.0      \\ \hline
Himmelblau        & 50                 & 654.0                 & 621.0                 & 621.0      \\ \hline
Schwefel        & 50                 & 639.0                 & 636.0                 & 636.0      \\ \hline
\end{tabular} 
\label{results-haea}
\end{table}

The results shown in \ref{results-ga} and in \ref{results-haea} confirm the previous intuition. The proposed approach is not statistically different \textit{only} in some functions that have lower dimensionality and simpler analytic form (with the exception of the Schewfel function). In the other functions, the proposed approach gives a better performance and in some functions the algorithm did not converge in the 499th iteration, with more fitness evaluations is expected to obtain even better results. 

\subsection{Analysis of operators behaviour}

\begin{figure}
	\centering 
	\includegraphics[scale=0.21]{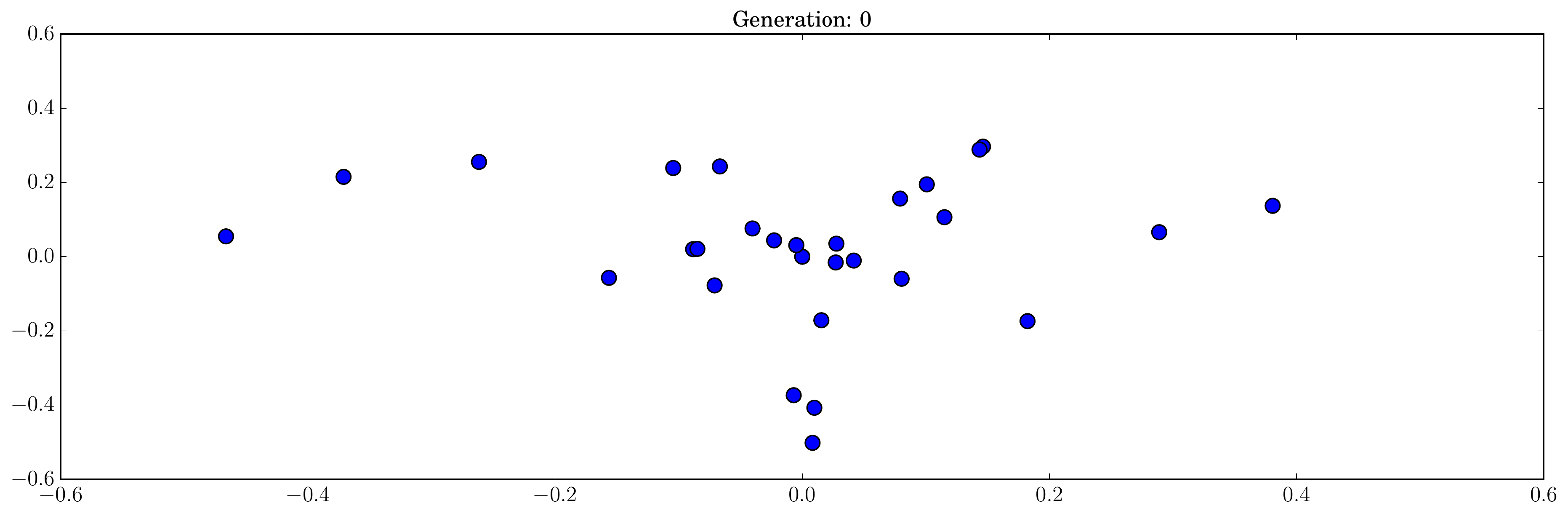} \hfill
	\includegraphics[scale=0.21]{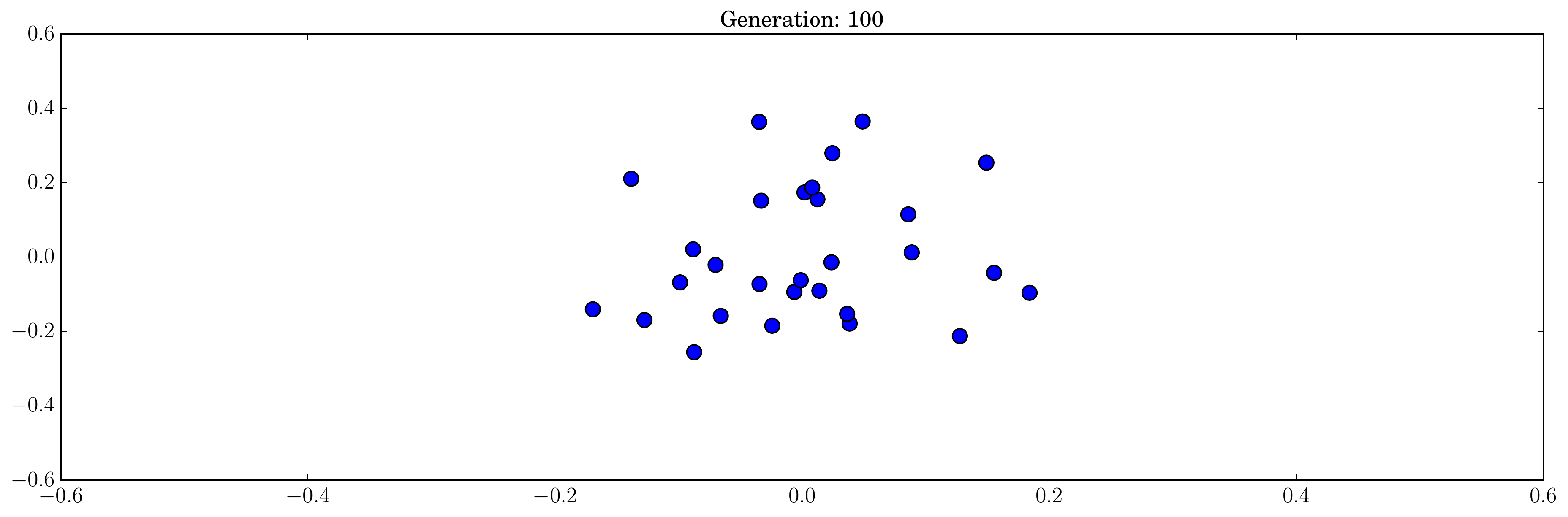} \hfill
	\includegraphics[scale=0.21]{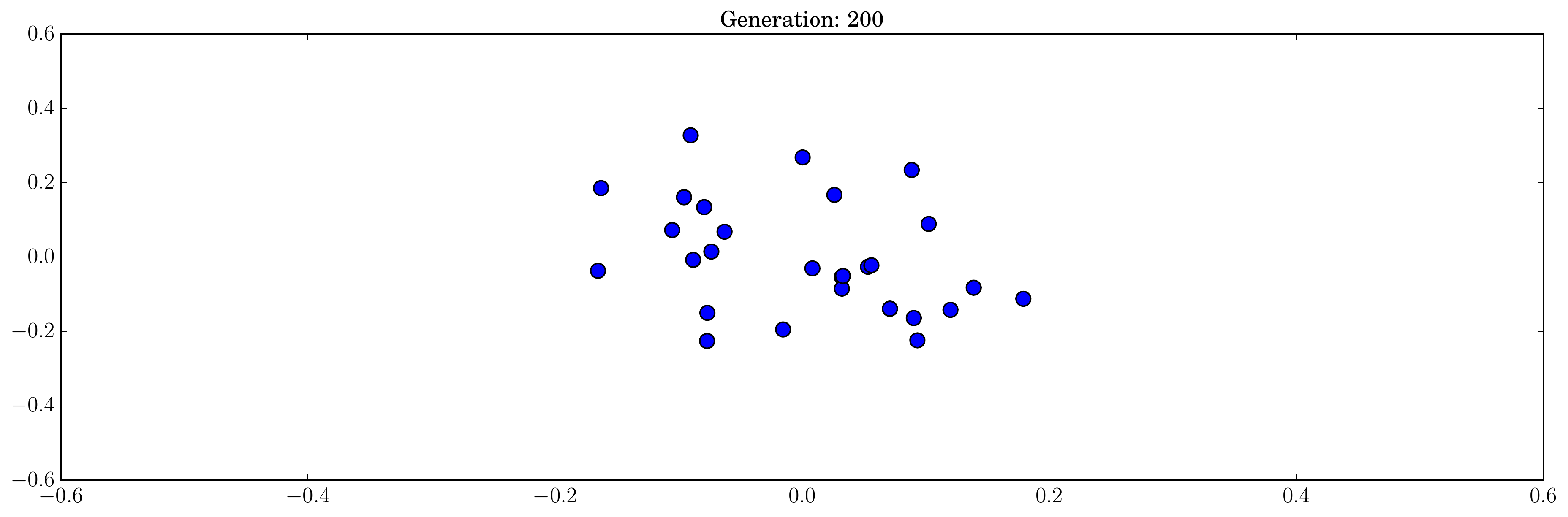} \hfill
	\includegraphics[scale=0.21]{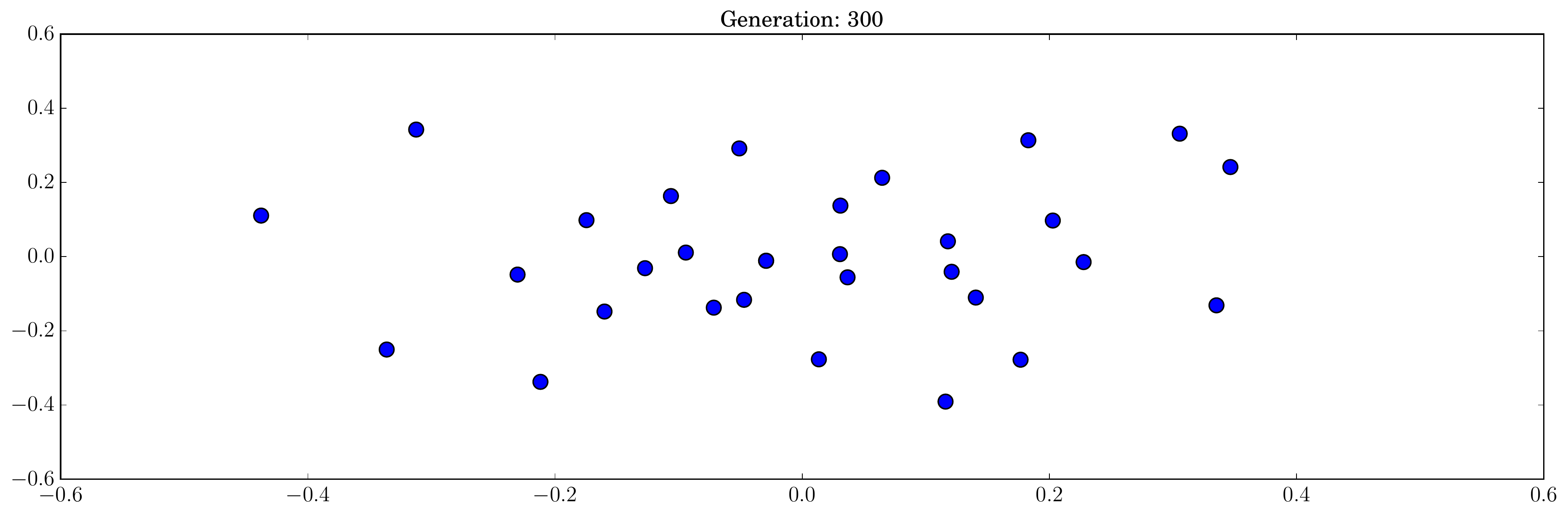} \hfill
	\includegraphics[scale=0.21]{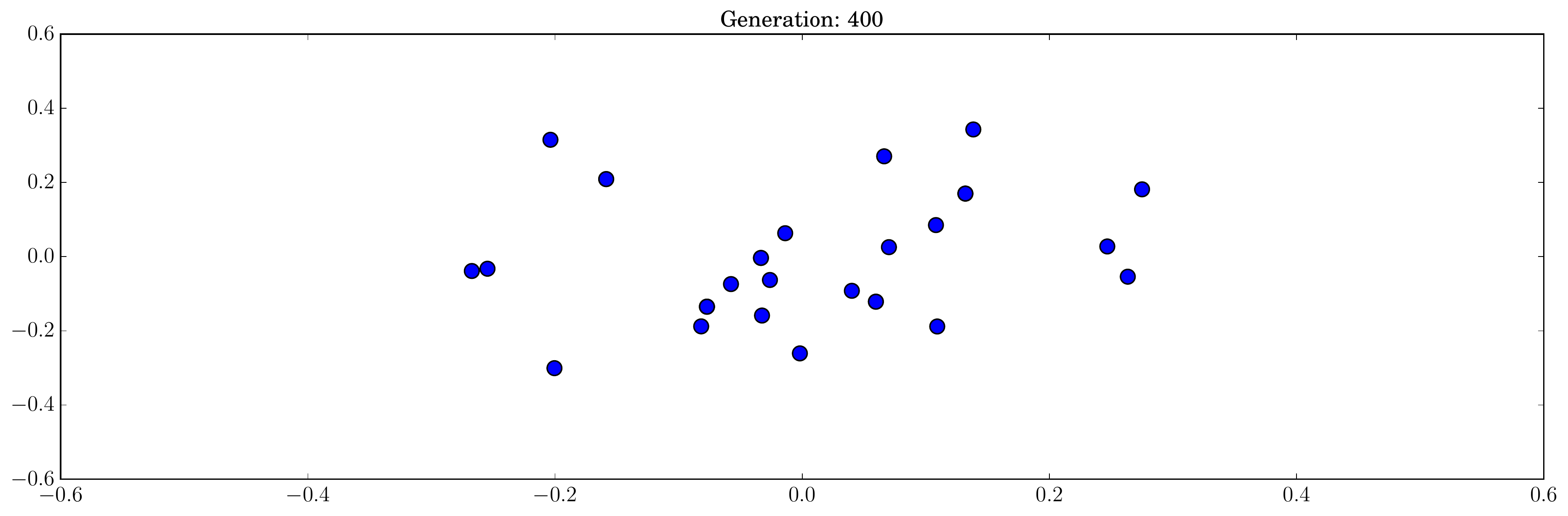} \hfill
	\includegraphics[scale=0.21]{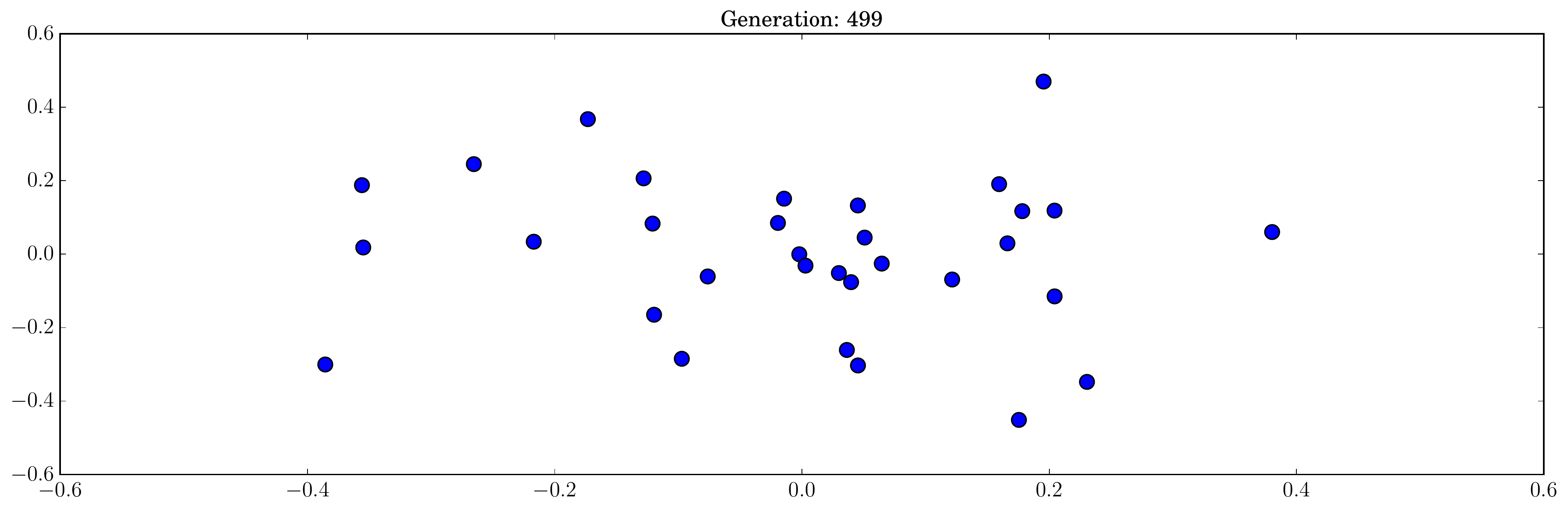} \hfill				
	\caption{Trees embedded in a 2D space in generation from 0 to 499. The operators are evolved to minimize the Ackley function. The magnitude of the coordinates is a result of the Multidimensional Scaling process which enforces the points to be close if their tree distance is small. The 2D embedding was computed using the sklearn \cite{scikit-learn} implementation of the SMACOF (Scaling by Majorizing a Complicated Function) algorithm \cite{de2011multidimensional}.}
	\label{generations}	
\end{figure}

{\bf Trees similarity}: The trees are stored per generation and are compared pairwise using the tree edit distance proposed by Zhang and Shasha \cite{treedistance}, this distance is the minimum number of operations to transform a tree into another tree. On each generation, the trees are transformed into a two dimensional space for visualization using multidimensional scaling with a the matrix of (normalized) pairwise distances. The graphical results are presented in figure \ref{generations}. The trees tend to converge to a single cluster, but they never group into a single one. The behaviour is to converge until certain point then start to separate and then group again and so on. In the last plot of figure \ref{generations} there is a snapshot of the 499th generation where the operators have not converged yet.

\begin{figure}

	\begin{subfigure}{\linewidth}
	\includegraphics[scale=0.3]{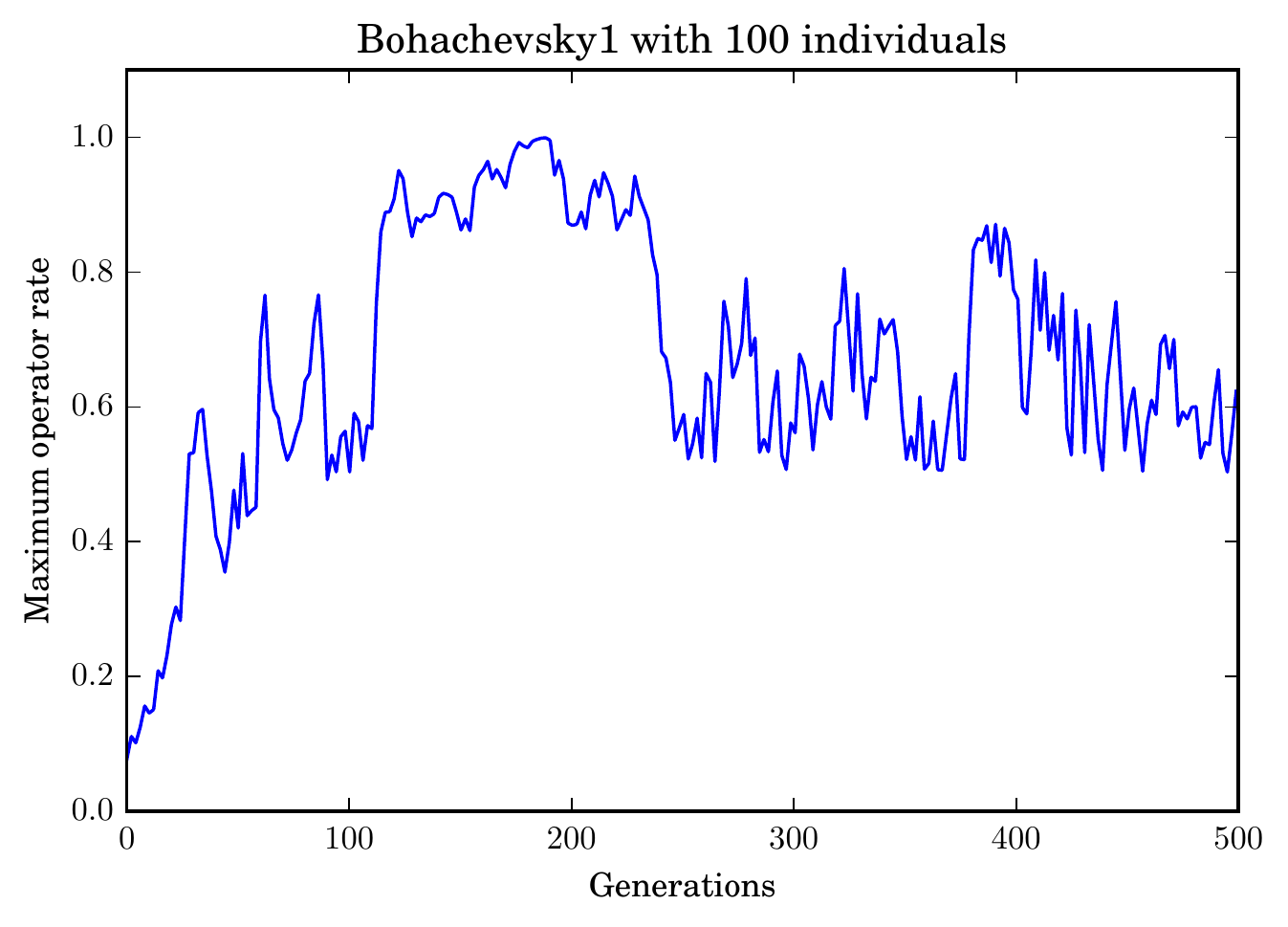}\hfill
	\includegraphics[scale=0.3]{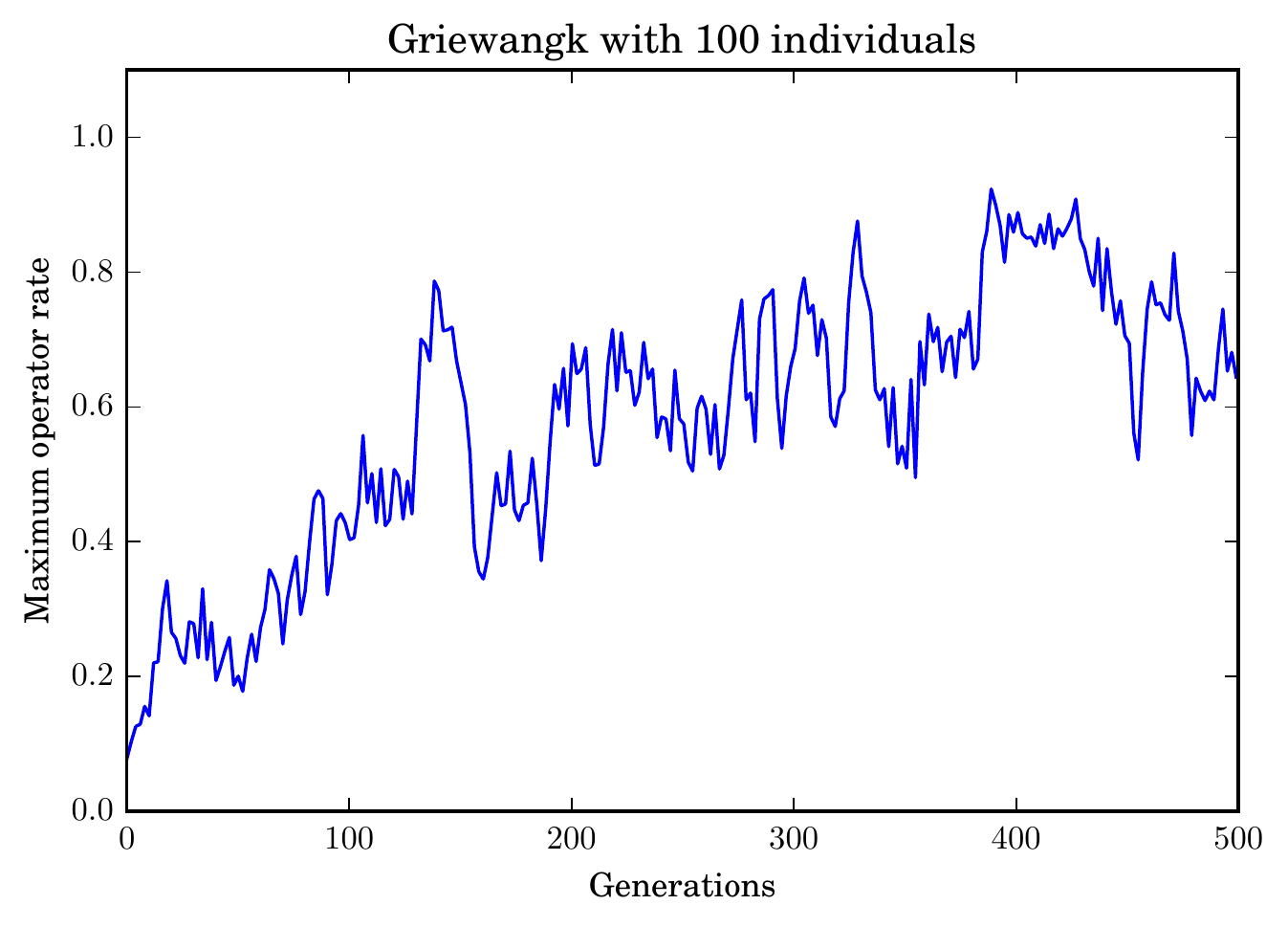}
	\caption{Best rates of Bohachevsky, and Griewangk function with 100 individuals in the population.}
	\end{subfigure}\par\medskip
	
	\begin{subfigure}{\linewidth}
	\includegraphics[scale=0.3]{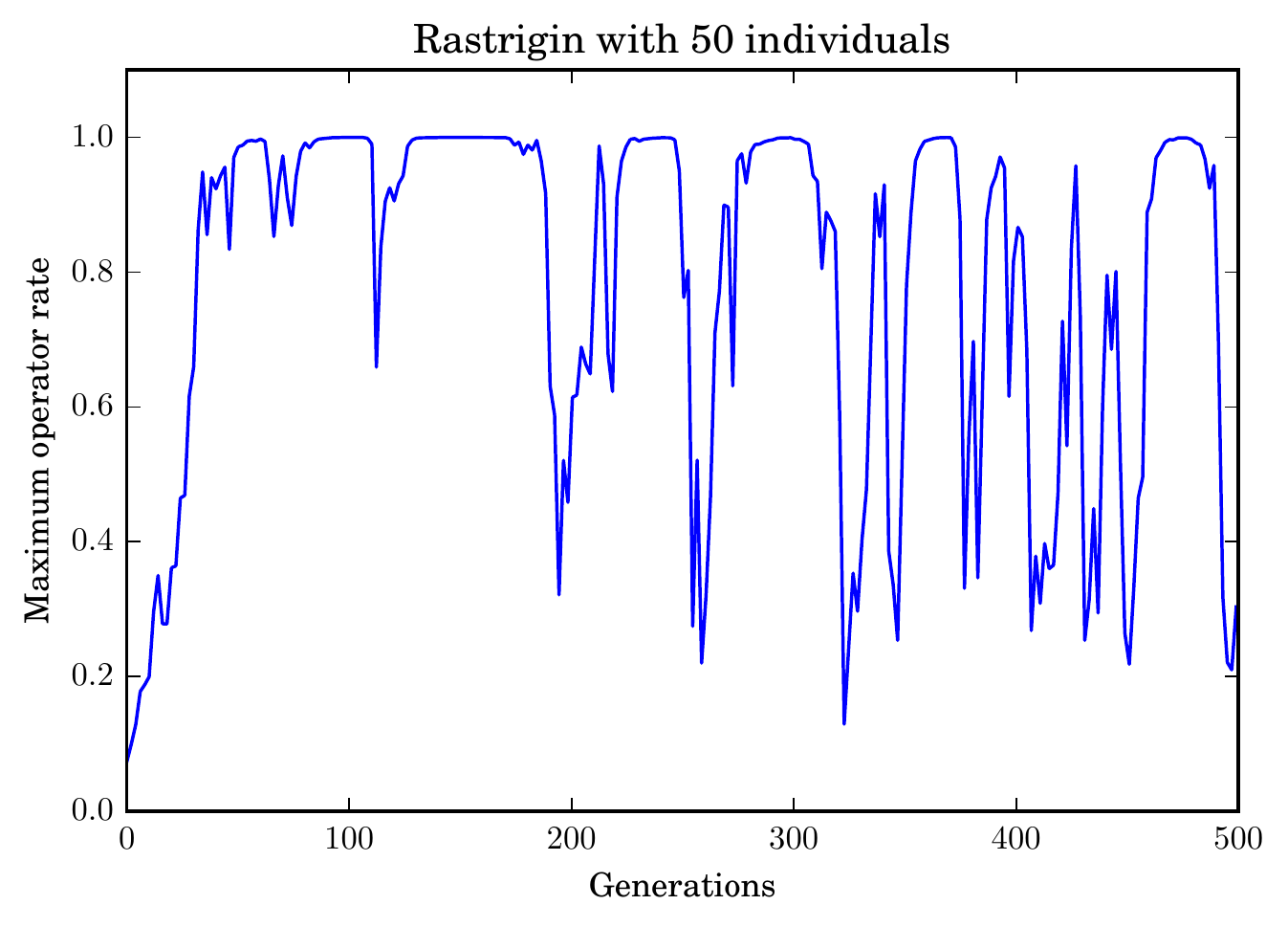}\hfill
	\includegraphics[scale=0.3]{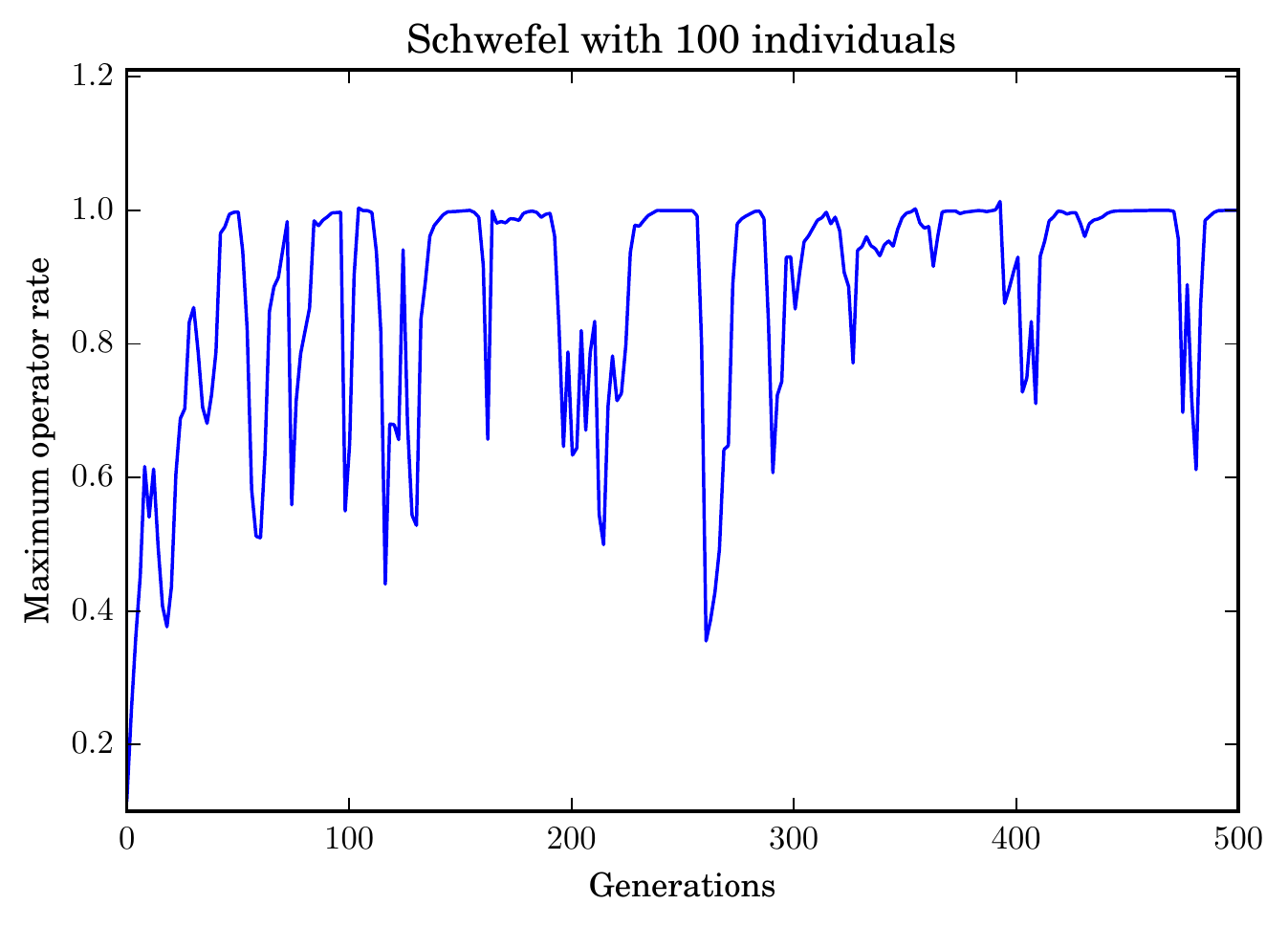}\hfill
	\caption{Best rates of Rastrigin, Schwefel, function with 50 and 100 individuals in the population.}
	\end{subfigure}\par\medskip
	\caption{Maximum rates over operators population}
	\label{maxrates}
\end{figure}

The trees never converge into a very similar tree, which is good because diversity is maintained, and it implies that different changes are applied to the candidate solutions, giving them the chance to delay the convergence but still produce good results. Another interesting result is the behaviour of the operator rates, where, once again, the dynamics change according to the ``hardness" of the tested function. Figure \ref{maxrates} shows the maximum rate (probability) of the operators population per generation. 

From figure \ref{maxrates} it can be seen, that the best rates per generation do not converge right away but ``oscillate" through the iterations of the algorithm. Whenever a set of rates is very high, the complement is going to have a very low probability to be selected because the selection method is proportional to the rates. The fact that the rates do not converge to a rate close to 1.0 means that operators that are applied to individuals are not always the same. Furthermore, the rates are also used to evolve operators, which contributes to maintain the overall diversity as was shown in the 2D embedded visualizations. These are empirical conclusions given the patterns revealed by the data, more rigorous analysis using statistical tools is out of the scope of this work.

\section{Conclusions and future work}

The proposed algorithm shows a fast convergence in most of the functions tested, and does not fall into premature convergence due to the generated diversity by the operators scheme and the punish/reward update, which puts enough pressure to achieve desirable results. Moreover, this scheme is easily applied to other kinds of problems without having to specify complex operators to generate new solutions, but only defining small atomic operators (which incorporate knowledge of the problem domain) and let the evolutionary process combine them. Finally, in numeric optimization problems there is no previous knowledge required about the function like gradients or the specific function, but in order to obtain better results and fast convergence it is useful to know the constraints for each dimension in order to maintain feasible solutions along the algorithm. It should be noted that there is an runtime overhead on the operators evaluation, as well on its selection according to its quality measure. However, this overhead remains constant with respect to the number of fitness evaluations, which is usually the bottleneck on real-world applications. Further more, in our experiments GA and HAEA converged quickly to bad local minima. Due to the selection pressure it can be hypothesized that with more computational resources they will not evolve better solutions because the diversity is greatly reduced. 

Future work includes applying this approach in other problems outside of the numerical optimization domain and possibly in other contexts like open ended evolution, non stationary functions, and multi-objective optimization, where self-adaptation in the breeding operators is needed in order to maintain genetic diversity. As usual with evolutionary strategies, the population size has a crucial role on maintaining diversity. Future work, also includes finding a way to self-adapt the population size with techniques related to the proposed approach, as well as applying this approach with separately evolved, small operator populations for every candidate solution that exchange information between each other.

Finally, we hope to do a more rigorous analysis of the operators convergence using appropriate statistical tests and more refined techniques to compare the trees structure and its relation with the problem being optimized.

\mbox{}

\bibliographystyle{splncs03}
\bibliography{bib}

\end{document}